\documentclass[10pt,letterpaper]{article}

\usepackage{cogsci}
\usepackage{xcolor}

\usepackage{graphicx}
\usepackage{amsmath,amssymb}
\usepackage{booktabs}
\usepackage{array}
\usepackage{caption}
\usepackage[colorlinks=true,citecolor=black]{hyperref}

\cogscifinalcopy % Uncomment this line for the final submission 

\usepackage{pslatex}
\usepackage{apacite}
\usepackage{float} % Roger Levy added this and changed figure/table
\title{Minding the Politeness Gap in Cross-cultural Communication}

 \author{Yuka Machino$^1$, Matthias Hofer$^1$, Max Siegel$^1$, Joshua B. Tenenbaum$^1$, and Robert D. Hawkins$^2$
 \\
$^1$Department of Brain and Cognitive Science, MIT \\ 
$^2$Department of Linguistics, Stanford University}

\begin{document}

\maketitle

\begin{abstract}
Misunderstandings in cross-cultural communication often arise from subtle differences in interpretation, but it is unclear whether these differences arise from the literal meanings assigned to words or from more general pragmatic factors such as norms around politeness and brevity.  
In this paper, we report three experiments examining  how speakers of British and American English interpret intensifiers like ``quite'' and ``very.''
To better understand these cross-cultural differences, we developed a computational cognitive model where listeners recursively reason about speakers who balance informativity, politeness, and utterance cost. 
Our model comparisons suggested that cross-cultural differences in intensifier interpretation stem from a combination of (1) different literal meanings, (2) different weights on utterance cost. 
These findings challenge accounts based purely on semantic variation or politeness norms, demonstrating that cross-cultural differences in interpretation emerge from an intricate interplay between the two.
\end{abstract}

\section{Introduction}
A central maxim of cooperative communication is that speakers aim to be \textit{informative}---providing relevant and useful information to their listeners \cite{Grice1975LogicAC}. 
However, speakers must balance many other considerations when choosing what to say, including politeness, efficiency, and clarity.
These alternative considerations may come into tension with informativity, giving rise to indirectness, vagueness, hedging, and other forms of pragmatic ambiguity that listeners must resolve. 
For example, when interpreting the strength of an utterance containing intensifiers like \emph{very} or \emph{kind of,} listeners may wonder whether the speaker is literally trying to convey a stronger or weaker value on the given scale, or whether they're just trying to convey a sense of politeness.
Consider the phrase ``That was very helpful of you.'' 
Does the speaker mean the help was exceptionally good, or are they simply being gracious? Similarly, ``I'm kind of tired'' could indicate mild fatigue or serve as a polite understatement \cite{brown1987politeness}.

These inferences are complex enough within a single cultural context, but they pose particular challenges when speakers and listeners come from different cultural backgrounds \cite{Goddard2012EarlyII,HaughBousfield12,thomas1983cross}. 
Corpus studies have extensively documented variation in how speakers use and interpret modifiers  \cite{Su2016, Stratton_2021,Romero2012ThisIS}.
For instance, the intensifier \emph{quite} is known to have different meanings in British and American English, functioning as an amplifier in American English but often as a downtoner in British English \cite{fowler2015fowler,desagulier2014visualizing}.
These differences extend beyond individual lexical items to broader patterns of how modifiers function in discourse--as devices for hedging, emphasis, or managing interpersonal dynamics \cite{Ruzaitė+2007, HAUGH20121017,SCHNEIDER20121022, WATERS20121051}. 
However, identifying the underlying source of these differences presents a challenge: what appears to be a fixed semantic difference could actually arise from variation in pragmatic factors. 
For example, if British speakers expect stronger politeness norms, they might discount downtoners like \emph{slightly} as polite hedging, leading to systematic interpretation differences even if the modifiers have the same meaning.

Recent work in computational pragmatics \shortcite{Yoon2018PoliteSE,lumer2022modeling} offers a promising framework to disentangle these factors.
The Rational Speech Act (RSA) framework models speakers as agents who maximize a utility function weighing epistemic goals against other social goals. Listeners, in turn, must reason about speakers' motivations to correctly interpret what is said. 
Crucially, this framework allows us to model the interface between the literal semantic meanings of modifiers and the pragmatic reasoning processes that operate over them. 
While this framework has already provided valuable insights into the pragmatics of polite language use within a single culture, it has not yet been extended to explain cross-cultural variation \cite{GoddardWierzbicka,goddard2013words,white2024communicate}.

\label{exp:dialogue}

\begin{figure*}[t!]
    \centering
    \includegraphics[width=\textwidth]{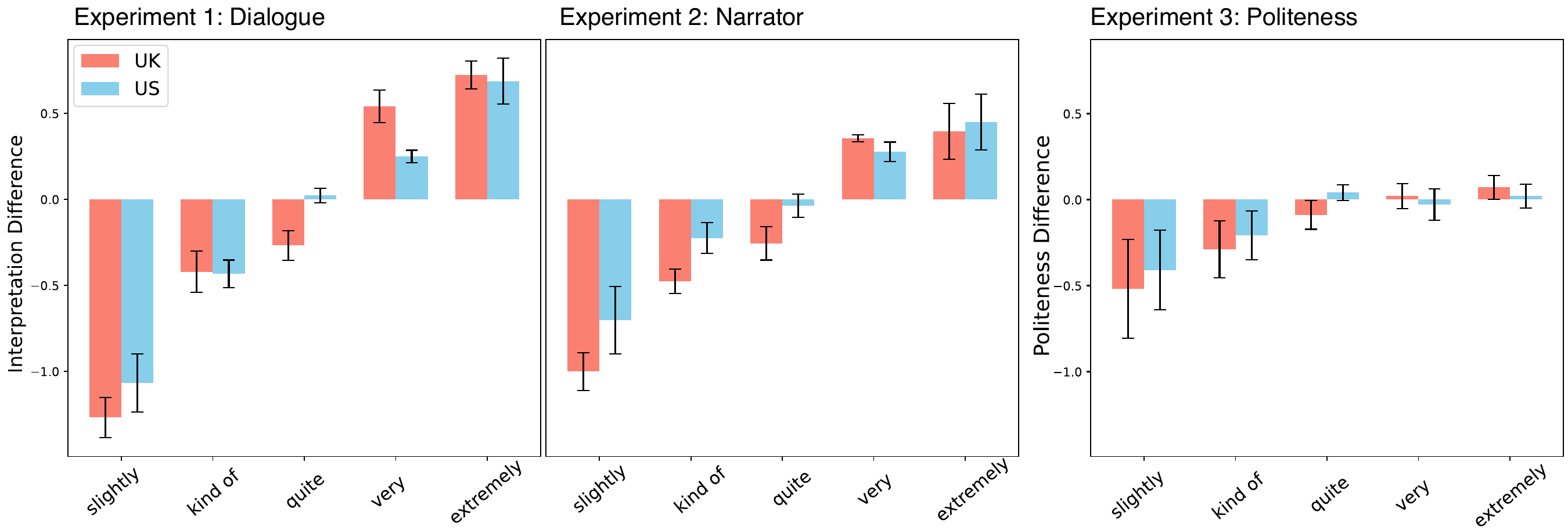}
    \caption{Results from three experiments reveal robust interpretation differences between UK and US participants. Experiments 1 and 2 measured interpretation differences embedded in different conversational contexts, while Experiment 3 measured perceived politeness. Error bars represent bootstrapped 95\% confidence intervals.}
    \label{fig:interpretation_differences}
\end{figure*}

In this study, we extend the RSA framework to account for cross-cultural variation in modifier interpretation between British and American English. Through three experiments, we test whether interpretation differences arise from semantic factors (literal meanings) or pragmatic factors (politeness norms). We then develop a computational model building on \citeA{Yoon2018PoliteSE} to quantitatively disentangle these contributions. By fitting culture-specific parameters for literal meanings, utility weights, and utterance costs, we find that variation in modifier interpretation arises through both semantic variation (e.g., different thresholds for \emph{quite} and \emph{very}) as well as pragmatic variation (e.g., the markedness of modifier production). These findings reveal that cross-cultural variation in interpretation emerges from a complex interplay of semantic and pragmatic factors, demonstrating the need to incorporate cultural variation into computational models of language understanding.

\section{Behavioral Experiments}
We conducted three experiments, each designed to isolate distinct explanations for differences in intensifier interpretation between English speakers in the US and UK. 
By systematically varying conversational context and speaker intent, our experiments  aimed to disentangle the relative contributions of semantic and pragmatic factors.

\subsection{Experiment 1: Dialogue Context}

We begin by establishing whether British and American English speakers systematically differ in how they interpret modifiers, providing a baseline measure of cross-cultural variation.

\subsubsection{Participants}

We recruited 49 participants via Prolific (23 from the UK, 26 from the US) to interpret a series of utterances with and without intensifiers. 

\subsubsection{Stimuli \& Design}
We crossed seven predicates (``exhausted'', ``boring'', ``difficult'', ``concerned'', ``understandable''. ``impressive'', ``helpful'') with five modifiers (\emph{slightly}, \emph{kind of}, \emph{quite}, \emph{very}, \emph{extremely}) in a within-subject design. 
Each predicate appeared in both an \textsc{Unmodified} version (e.g. ``You've been helpful'') and a \textsc{Modified} version (e.g., ``You've been very helpful''), embedded in short vignettes designed to vary in non-linguistic social features such as speaker-listener relationships, social hierarchy, and formality of interaction, controlling for contextual factors (see Table~\ref{tab:scenario_versions}). 
Participants rated each utterance on a continuous scale from ``minimally {predicate}" to ``maximally {predicate}". 
Each participant completed 70 randomized trials: 35 modified and 35 unmodified baseline sentences, enabling within-subject comparison of modifier effects.

\begin{table}[b!]
\centering
\small
\caption{Scenario variants and corresponding questions.}
\label{tab:scenario_versions}
\begin{tabular}{@{}p{0.16\linewidth}p{0.56\linewidth}p{0.2\linewidth}@{}}
\toprule
\textbf{Version} & \textbf{Scenario} & \textbf{Question} \\ \midrule
\textbf{Exp.1: Dialogue} & Amy asks her friend, Lisa, for some tips on baking bread: \textit{"Thank you for those tips. You've been very helpful."} & How helpful did Amy find Lisa's tips? \\ \addlinespace
\textbf{Exp.2: Narrator} & Amy asks her friend, Lisa, for some tips on baking bread. The tips are very helpful, and Amy thanks Lisa for it. & How helpful did Amy find Lisa's tips? \\ \addlinespace
\textbf{Exp.3: Politeness} & Amy asks her friend, Lisa, for some tips on baking bread: \textit{"Thank you for those tips. You've been very helpful."} & How polite is Amy being? \\
\end{tabular}
\end{table}

\subsubsection{Results}
To determine whether British and American participants systematically differed in their modifier interpretations, we fit a linear mixed-effects model with random effects for participant and scenario. We predicted the within-participant difference between the {\sc baseline} and {\sc modified} conditions for each scenario.
To control for individual differences in scale use, we first $z$-scored all responses within each participant. 
We then computed a modifier effect score by subtracting the $z$-scored response of the unmodified version from the $z$-scored response of the modified version of each scenario, thereby isolating the effect of the modifier.
We compared a simpler model including only fixed effects of country (US vs. UK) and modifier against a more complex model including a country by modifier interaction term. 
A nested likelihood ratio test revealed that the model allowing for different effects for each modifier provided a significantly better fit ($\chi^2(4) = 14.48, p = .006$), confirming systematic cross-cultural differences in modifier interpretation.

Post-hoc comparisons for individual modifiers (see leftmost graph in \autoref{fig:interpretation_differences}) revealed that these differences were specific rather than general: \emph{quite} was interpreted as more of a downtoner by British participants compared to Americans ($\beta = -0.28, SE = 0.12, p = .023$); \emph{very} was interpreted as a stronger amplifier by British participants ($\beta = 0.28, SE = 0.13, p = .032$), while no reliable differences emerged for the rest of the modifiers. Across both groups, the modifiers followed a consistent strength hierarchy: \emph{slightly} $<$ \emph{kind of} $<$ \emph{quite} $<$ \emph{very} $<$ \emph{extremely} (all pairwise comparisons $p < .001$).

To further contextualize these findings, we tested whether modifier effects varied by predicate valence. We categorized predicates as positive (understandable, impressive, helpful) or negative (exhausted, boring, difficult, concerned) based on their typical evaluative connotations. 
A mixed-effects model including the additional effect of valence suggested that modifier interpretation varied systematically with valence ($\chi^2(4) = 20.83, p = .0003$), particularly for downtoners: both \emph{slightly} ($p = .017$) and \emph{kind of} ($p = .042$) showed stronger weakening effects with positive predicates compared to negative ones.
British participants' weaker interpretation of \emph{quite} occurred only for negative predicates ($p = .022$), while their stronger interpretation of \emph{very} was most pronounced with positive predicates ($p = .053$).

\subsection{Experiment 2: Narrator Context}\label{exp:narrator}

Our findings in Experiment 1 support the observation that intensifiers have different effects across US and UK cultures, but leave open whether these differences stem from semantic or pragmatic factors. 
These scenarios embedded utterances in a dialogue, where social and politeness considerations naturally arise.
In Experiment 2, we designed variants of the same scenarios to minimize explicit politeness considerations by re-framing the final sentence as an omniscient narrator's description, akin to an inner monologue (see \autoref{tab:scenario_versions}). 
In principle, this manipulation should remove the social pressure to be polite. Without a conversational partner, there is no overt reason to hedge, intensify, or soften statements for politeness. 
Thus, if the differences in modifier interpretation we observed between US and UK participants in Experiment 1 persist in a non-communicative setting, it would suggest that these differences may be more due to differences in the literal meaning of intensifiers. 
However, if the differences disappear or significantly reduce, this would suggest that pragmatic considerations -- such as politeness norms -- play a more significant role in shaping cross-cultural differences.
 
\subsubsection{Results}
We recruited 39 participants, 19 from the UK and 20 from the US; aside from using the modified variants of the stimuli, all other elements of the design were the same.
As in Experiment 1, we examined difference scores for each modifier. 
To quantify how much of the differences in Experiment 1 still persist in the absence of the need of the speaker to be polite, we added per-item averaged responses from Experiment 2 as an additional predictor in the same linear mixed-effects model we used for Experiment 1. 
Including the effect of framing (dialogue vs. narrator) 
significantly improved model fit in a likelihood ratio test ($\chi^2(1) = 7.31, p = .007$),  indicating some effect of the narrator manipulation. Critically, we still observed persistent differences in modifier interpretation between UK and US participants (see \autoref{fig:interpretation_differences}, middle), suggesting that perceived politeness pressures alone do not fully explain the interpretation differences found in Experiment 1. 
Notably, while the cross-cultural gap for \emph{very} closed, the modifier \emph{quite} still showed significant variation.

Our results also suggested that participants consider pragmatic factors: if the only difference between the narrator and dialogue scenario were politeness, we would expect \emph{stronger} interpretations of modifiers in the narrator condition, since modifiers should now purely convey information without ambiguity about social constraints. 
If anything, we found marginal evidence suggesting the opposite effect ($p=0.13$), that modifiers actually influenced interpretation less in the narrator condition, meaning that listeners inferred additional meaning from overt speech compared to inner monologue.
One possible explanation is utterance cost: in a communicative setting, the mere act of choosing to modify an utterance may signal additional meaning beyond the literal semantics of the intensifier. In contrast, in a (non-communicative) narrator setting where there is no overt utterance production, such cost-based inferences may not apply, leading to weaker modifier interpretations. Importantly, the extent of this reduction varied across cultures, suggesting that the perceived ``cost" of an utterance, and the inferences listeners draw from modification may be culturally specific.

In summary, Experiment 2 provides indirect evidence that at least some of the observed variation is due to different literal meanings between the UK and the US. However, we could not fully rule out pragmatic influences at play in the narrator condition itself. As we have observed with the reduction of modifier effect in the narrator condition, multiple pragmatic factors changed between the narrator and dialogue condition. In addition, participants may have implicitly interpreted the narrated inner monologue as a kind of speech act, introducing residual pragmatic considerations into their judgments. To address this, Experiment 3 directly measured perceived politeness for each predicate-modifier combination, allowing us to explicitly quantify the role of politeness in modifier interpretation.

\subsection{Experiment 3: Politeness Ratings}\label{exp:politeness}
% While Experiment 2 provided indirect evidence that pragmatic politeness considerations influence modifier interpretation, it did not completely rule out pragmatic influences on the narrator condition itself. 
To explicitly quantify the role of politeness norms, Experiment 3 directly measured perceived politeness for the same predicate-modifier combinations used in Experiment 1. 

\subsubsection{Results}

We recruited 40 participants from the UK and US, ensuring that each participant saw the same set of scenarios as in previous experiments.
Instead of asking participants to interpret the meaning, we asked them to rate the politeness of each utterance. 
The results partially support our hypothesis, that intensifying a negative predicate is perceived as impolite and intensifying a positive predicate is polite, whereas downtoning has the opposite effect. Interestingly the modifiers which exhibited the highest sensitivity to valence were \emph{kind of} and \emph{slightly}, exactly the same as in Experiment 1. For both of these modifiers, modifying (downtoning) positive predicates were perceived as impolite, $p<0.01$. The other modifier-valence pairs did not have a significant effect on politeness.

In order to compare main effects of the modifiers, we flipped the sign of the politeness difference according to the valence (see \autoref{fig:modelVsData}, right for the mean and standard error across predicates).  Intuitively, the more negative this value is, the more polite they are in negative contexts and impolite in positive contexts, hence exhibiting more downtoner like characteristics according to our hypothesis. We found significant variation between modifiers, and the order generally followed this predicted trend, supprting the view that the change in degree and the politeness interpretation are highly related (the modifiers \emph{slightly} $<$ \emph{kind of} $<$ \emph{quite} were both statistically significant with $p< 0.01$, while comparisons between \emph{quite}, \emph{very}, and \emph{extremely} did not reach statistical significance).

To test whether cross-cultural differences in modifier interpretation are driven by politeness-based pragmatic inferences, we fit a linear mixed-effects model including perceived politeness judgments as a predictor. Politeness was included as an per-item averaged additional measure as predictor into the linear model.
We found that incorporating perceived politeness as a predictor significantly improved the fit of the dialogue data, $\chi^2(1)$ = 27.90, $p < 0.001$, suggesting that politeness perceptions capture relevant contextual information which affect modifier interpretation. Furthermore, while politeness ratings were highly correlated across UK and US participants, we observed systematic differences between the two groups. In particular, modifiers that showed interpretation differences in the narrator condition (e.g., \emph{quite}) also showed differences in politeness perception, suggesting that some semantic differences between UK and US English contribute to the effects of its role as modifiers. 

To test whether difference in politeness perception explains cross-cultural differences in modifier interpretations, we fit a model which included modifier-predicate interactions (and hence encoded valence and broader contextual information), and investigated whether adding the country dependent per-item averaged politeness perception increased fit to the modifier interpretations from Experiment 1. We found that the improvement was not significant ($p=0.60$), suggesting that cross-cultural differences in politeness perception does not linearly explain interpretation differences.
Taken together, while these findings provide strong evidence that politeness norms systematically influence modifier interpretation, the difference in politeness norms as measured in Experiment 3 did not directly translate into differences in pragmatic interpretation of modifiers across cultures.

\section{Computational Model}
\label{CompModel}

Our behavioral experiments established two clear patterns: British and American speakers interpret certain modifiers differently, and these differences are not fully explained by differences in politeness perception.
However, culture-dependent politeness norms may affect utterance interpretation in more nuanced ways. 
In order to quantitatively investigate how semantic and pragmatic differences contribute to interpretation differences between cultures, we build on the Rational Speech Act (RSA) model of politeness proposed by \citeA{Yoon2018PoliteSE}. 
In this framework, speakers choose utterances to optimize multiple goals simultaneously -- being informative, being kind, and being concise. Listeners then work backwards from the utterance to infer both what the speaker meant and why they chose that utterance as opposed to another one. 
Crucially, we extend this framework to allow cross-cultural variation in three key components: (1) the literal meanings assigned to modifiers, (2) the relative importance of politeness versus informativity, and (3) the additional cost of adding modifiers.

\subsubsection{Modeling framework}

Following \shortciteA{Yoon2018PoliteSE} and \citeA{lumer2022modeling}, we formalize speaker preferences in a utility that balances three competing objectives: \textit{informativity}, \textit{conciseness}, and \textit{kindness}.
$$U_{S}(w|s,\phi_{i},\phi_{s}) = \phi_{i}\cdot U_{i}(w|s)+\phi_{s}\cdot U_{s}(w)-C(w)$$
where $w$ is the speaker's utterance (``very helpful''), $s$ is the true underlying state that the speaker wants to communicate (i.e. the degree of helpfulness), and $\phi_{i}, \phi_s$ are weights controlling the importance of being informative and kind, respectively.

We define the informativity term to be the log-likelihood that a literal listener $L_0$ correctly identifies the intended state: $U_{i}(w|s) = \ln(P_{L_0}(s|w))$, where the literal listener interprets utterances based on their  literal semantic denotation, $$P_{L_0}(s|w)\propto[\![w]\!]_{\theta}(s)\cdot P(s),$$ where $[\![w]\!]_{\theta}$ is a parameterized function evaluating to 1 if the utterance $w$ is compatible with the state $s$, and $\epsilon<<1$ otherwise.
The social utility term $U_{s}(w)$ represents the perceived kindness or social appropriateness of the utterance: for the purposes of this paper, we define this term directly through the ratings collected in Experiment 3.
Finally, we define the utterance cost, $C(w)$ as the cost required to produce an utterance, penalizing the production of additional modifiers utterances.
The key insight is that when speakers add modifiers, listeners infer there must be a reason to depart from the simpler unmodified form. 

Using this utility, we model the speaker's utterance probability as the softmax of the combined utility marginalized over each state, i.e.
$$P_{S_1}(w|s,\phi_{i},\phi_{s}) \propto \exp[U_{S_1}(w|s,\phi_{i},\phi_{s})]$$
A pragmatic listener who knows the speaker is optimizing this utility can then infer the likely state by inverting the speaker's production model:
$$P_{L_1}(s|w,\phi_{i},\phi_{s}) \propto P_{S_1}(w|s,\phi_{i},\phi_{s})\cdot P(s).$$

\subsubsection{Culturally dependent parameters}

To understand which aspects of this model vary across cultures, we compare models where different parameters are allowed to vary between British and American English speakers. 
We investigate two related hypotheses about how pragmatic factors may affect modifier interpretations. 
First, we hypothesize that differences in politeness perception across cultures explain differences in the interpretation of at least some modifiers.  
If this hypothesis holds, politeness perception will affect modifier interpretation, hence the best fitting model would have a non-zero coefficient on social utility $\phi_{s}$. 
Second, we hypothesize that differences in the way cultures trade off pressures to be polite, informative and concise lead to differences in interpretation. 
If the second hypothesis is true, we would expect to see different values for $\phi_{s},\phi_{i}$ and $C(w)$ between cultures.

Our full model has a total of fifteen parameters.
Twelve of these parameters control the literal semantics of each modifier. 
The literal meaning, which corresponds to the denotation function $[\![w]\!]$ in the model, uses a (smooth) double threshold, in which the function evaluates to 1 between the two thresholds, and 0 outside that interval. 
Therefore, the literal meaning of each modifier is described with two parameters (the lower and upper threshold), giving a total of ten parameters across all modifiers.
We further have two additional thresholds which captures the baseline range of values which are acceptable without any modifier, bringing us to a total of twelve semantic parameters.
The remaining three parameters capture possible differences in pragmatic factors: the utterance cost, $C(w)$, and the weight on social and informational utility $\phi = (\phi_{s},\phi_{i})$. 

\subsection{Results}
\subsubsection{Model comparison}
\begin{table}[t]
\centering
\caption{Results of model comparisons.}
\label{tab:ModelComparison}
\begin{tabular}{@{}p{0.07\linewidth}p{0.24\linewidth}p{0.06\linewidth}p{0.12\linewidth}p{0.13\linewidth}p{0.13\linewidth}@{}}
\toprule
\textbf{} & \textbf{Culturally Diff. Params} & \textbf{df} & \textbf{Log Loss} & \textbf{AIC} & \textbf{BIC}\\\midrule
\textbf{M1} & none &15 & 11250& 22530 & 22622\\ \addlinespace
\textbf{M2} & soc &16& 11250&22532& 22629\\ \addlinespace
\textbf{M3} & inf &16& 11244&22520&22618\\ \addlinespace
\textbf{M4} & cost &16& 11248& 22529 & 22627 \\ \addlinespace
\textbf{M5} & cost + inf &17& 11236&22507&22611 \\ \addlinespace
% \textbf{M6} & 'none' threshold& 17& 11250
% & 22533& 22637\\ \addlinespace
% \textbf{M7} & 'slightly' threshold& 17& 11250
% & 22533 & 22637\\ \addlinespace
% \textbf{M8} & 'kind of' threshold& 17& 11249
% & 22531 & 22635\\ \addlinespace
\textbf{M6} & 'quite' only& 17& 11226
& 22485 & \textbf{22590}\\ \addlinespace
\textbf{M7} & 'very' only& 17& 11227
& 22489 & 22593\\ \addlinespace
% \textbf{M11} & 'extremely' threshold& 17& 11249
% & 22532 & 22636\\ \addlinespace
\textbf{M8} & thresholds &27&  11219 &  22491& 22656\\ \addlinespace
\textbf{M9} & all & 30 & \textbf{11194} & \textbf{22448} & 22631\\ \addlinespace
\bottomrule
\end{tabular}
\end{table}

We evaluated our hypotheses through a systematic model comparison, allowing different subsets of parameters to vary between UK and US participants. 
These parameters include the semantic thresholds for each modifier's literal meaning as well as weights placed on informativity, social utility, and cost terms in the speaker utility. 
By comparing models where only semantic thresholds vary, where only pragmatic weights vary, or where the full combination varies, we can identify which types of cross-cultural variation are necessary and sufficient to explain the behavioral patterns in our data.

Results are shown in Table \ref{tab:ModelComparison}, using AIC and BIC as measures of model fit.
The best-fitting model (M9) integrated cross-cultural differences in both semantic and pragmatic factors, indicating that neither semantic nor pragmatic variation alone sufficiently explains the observed patterns. Indeed, all models that allowed variation in literal meaning thresholds (e.g. M8) showed substantial improvement over the baseline model (M1), demonstrating that literal semantic differences were necessary to capture the full pattern of responses. 
Notably, allowing just the 'quite' threshold to vary (M6) captures most of this improvement and achieves the best BIC score (which strongly penalizes additional parameters), consistent with our observation that 'quite' showed the largest cross-cultural difference.

While individual pragmatic parameters showed minimal improvement when varied alone (M2–M4), combining them yielded substantial gains. Model M5, which varied both cost and informativity weights, improved fit notably over models varying either parameter alone. This pattern indicates that cultural differences in the perceived cost of producing a modifier, combined with different weights on informativity, work together to shape utterance interpretation rather than operating independently.

\begin{figure}[t!]
    \centering
    \includegraphics[width=1\linewidth]{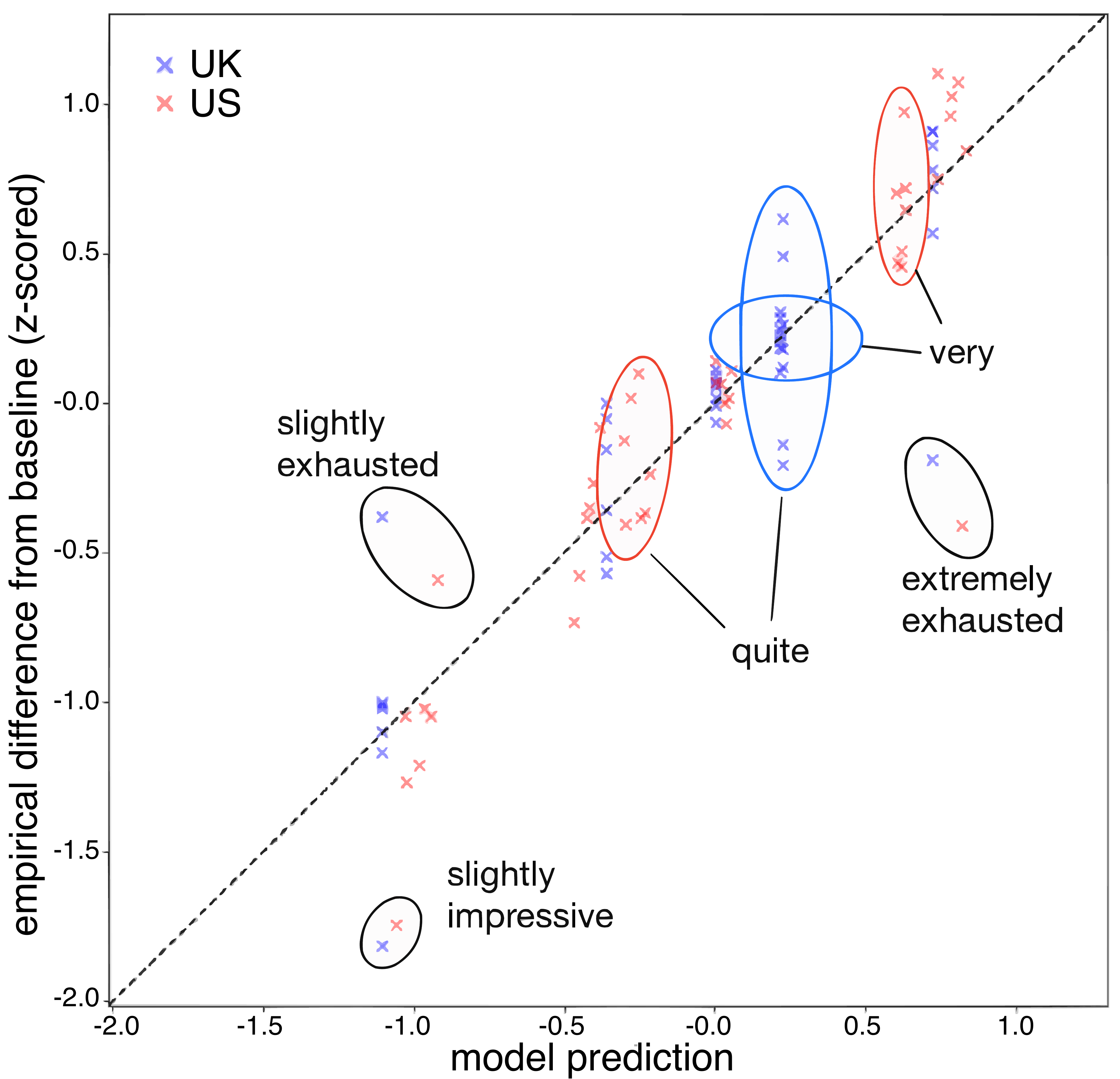}
    \caption{Comparison between model predictions and behavioral response. The predictions generated are from the model parameters specifically optimized to fit each country's data separately. Each cross represents a modifier-predicate pair.}
    \label{fig:modelVsData}
\end{figure}

\paragraph{Analysis of pragmatic factors}

To further validate the importance of pragmatic factors, we examined whether constraining specific parameters to be culture-invariant would degrade model fit. 
We minimally modified the best fitting models for the UK and US so that they share specific parameters while keeping others culturally dependent.
Forcing utterance cost, $C(w)$, or the weight on the informativity term, $\phi_{inf}$ to be the same across cultures, while keeping the other parameters the same as their respective best-fitting value, degraded the log loss significantly. 
These results further support the interpretation that pragmatic differences combine with semantic differences to play a significant role in shaping utterance interpretation.

Contrary to our initial expectations, the social utility term $U_{soc}$ did not play a significant role in any of these models. 
When we ignored the social utility component by setting the coefficient to zero, this only increased the log loss by at most two points, indicating that the additional complexity introduced by adding the politeness component in our model was not justified. This finding seemingly conflicts with our behavioral analyses, which suggested that politeness perceptions capture relevant contextual information for modifier interpretation.
This discrepancy suggests that while the politeness perceptions measured in Experiment 3 were relevant contextual information in predicting modifier interpretation, they did not affect interpretation in the way expected by our computational model. Our model assumes a direct trade-off between informativity and politeness, but the politeness ratings from Experiment 3 may capture other pragmatic factors such as sarcasm or exaggeration that are not accounted for in our current framework. \autoref{fig:modelVsData} suggests that despite this limitation, the best-fitting models for each culture successfully capture the overall pattern of behavioral responses, with model predictions closely tracking empirical data for most modifier-predicate combinations.

\section{Discussion}
This study proposed a novel approach to investigating how pragmatic communication varies between cultures by combining behavioral experiments with computational modeling. Through three experiments, we found systematic differences in modifier interpretation between British and American English speakers, revealing that these differences arise from both semantic and pragmatic factors. Our computational analysis provided quantitative evidence that cross-cultural variation emerges from the interaction of multiple factors: different literal meanings assigned to modifiers (particularly \emph{quite}), different weights placed on informativity, and different perceived markedness of modification. These findings challenge purely semantic or purely pragmatic accounts of cross-cultural variation, supporting instead an integrated view accounting for variation both in the literal meanings of expressions and the pragmatic reasoning processes that operate over them \cite{culpeper2017palgrave}.

Our model's surprising insensitivity to social utility in modifier interpretation reveals important limitations in current approaches to modeling politeness. Despite behavioral evidence that politeness perceptions influence interpretation, the $U_{soc}$ term contributed minimally to model fit. This limitation may suggest that the ratings from Experiment 3 were a poor proxy for social utility, or may reflect our assumption of uniform politeness norms within each culture. In reality, significant variation exists within both the UK and US due to regional dialects, socioeconomic factors, and diverse speech communities \shortcite{van2024sociopragmatic,schneider2024pragmatic}. By modeling social utility with a single culture-wide parameter, we may have obscured the role of politeness by averaging over heterogeneous subcommunities with different conventions. A hierarchical modeling approach that captures both within-culture and between-culture variation could better reveal how politeness shapes modifier interpretation \cite{troutman2022sassy,spencer2021intercultural}.

Further, the model's insensitivity to predicate variation is evident in \autoref{fig:modelVsData}, where model predictions tend to cluster by modifier type regardless of the predicate being modified. This insensitivity resulted in systematic deviations for certain modifier-predicate combinations, particularly infrequent or marked expressions. These outliers suggest that our model is failing to capture more fine-grained aspects of interpretation. Our framework assumes that modifiers operate uniformly across predicates, but in reality, the social implications of an utterance depend heavily on the specific predicate and conversational context \cite{sawada2017pragmatic}. For instance, saying ``I'm cold" might be perceived as merely informative in some contexts but as rude in others where it implies fault or imposes on the listener (e.g., suggesting they should close a window or turn up the heat). Similarly, ``extremely exhausted'' may be interpreted as hyperbolic precisely because ``exhausted'' already denotes an extreme state, making further intensification marked and potentially conveying sarcasm or complaint.

These observations point to the need for finer-grained representations that capture how different predicates interact with politeness considerations in context-specific ways. The politeness ratings collected in Experiment 3 likely reflect a complex mixture of factors including face management, social distance, imposition, and conventional implicatures that our model's architecture cannot disentangle. Future work might benefit from richer representations of social meaning that go beyond scalar utilities to capture these predicate-specific and context-dependent pragmatic enrichments \cite{sorensen2025value}.

Successful cross-cultural communication requires understanding not only different word meanings but also different conventions for interpreting them in context. Our finding that cultures assign different weights to informativity and utterance cost suggests speakers must recalibrate expectations about appropriate directness or elaboration when crossing cultural boundaries\cite{kecskes2023language}.
By quantifying the relative contributions of semantic and pragmatic factors, we can move toward more mechanistic explanations of how culture shapes communication. 
\vspace{2em}
\fbox{\parbox[b][][c]{\linewidth}{\centering {All code and materials available at: \\

\href{https://github.com/yukam997/PolitenessAcrossCultures}{\url{https://github.com/yukam997/PolitenessAcrossCultures}}

}}}

\section{Acknowledgments}
The authors thank the anonymous reviewers for their helpful feedback, as well as members of the Computational Cognitive Science Lab at MIT, and members of the Social Interaction Lab at Stanford for valuable discussions. We also thank Kartik Chandra for help implementing the computational model in memo.
Yuka Machino was supported by the Ezoe Recruit Memorial Foundation.
% In the \textbf{initial submission}, please \textbf{do not include
%   acknowledgements}, to preserve anonymity.  In the \textbf{final submission},
% place acknowledgments (including funding information) in a section \textbf{at
% the end of the paper}.

\bibliographystyle{apacite}

\setlength{\bibleftmargin}{.125in}
\setlength{\bibindent}{-\bibleftmargin}

\bibliography{CogSci_Template.bib}

\section{Appendix}

\subsubsection{Model Implementation}
We used the memo language \cite{chandra2025memo} to implement the computational model of the recursively reasoning listener.

\subsubsection{Optimization Process} We used the Covariance Matrix Adaptation Evolution Strategy \cite{hansen2023cmaevolutionstrategytutorial}  for parameter optimization as it converged more quickly and to lower values compared to its counterparts such as SGD or Adam.

\subsubsection{Modeling state prior as Gaussian}

\begin{figure}[t]
    \centering
    \includegraphics[width=1\linewidth]{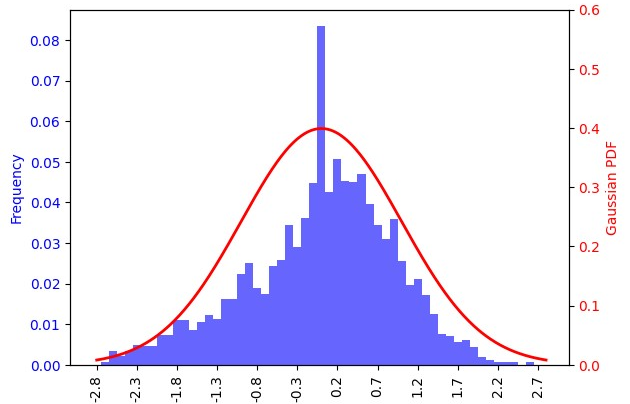}
    \caption{Gaussian state prior.}
    \label{fig:state_prior}
\end{figure}

We modeled the state prior as a normal distribution with mean 0 and variance 1. As seen in \autoref{fig:state_prior}, this assumption matches relatively well with the distribution of z-scored responses from participants suggesting that participants expect states to be distributed normally a priori.

\subsubsection{Double thresholds for modifier semantics} 
\begin{figure}
    \centering
    \includegraphics[width=\linewidth]{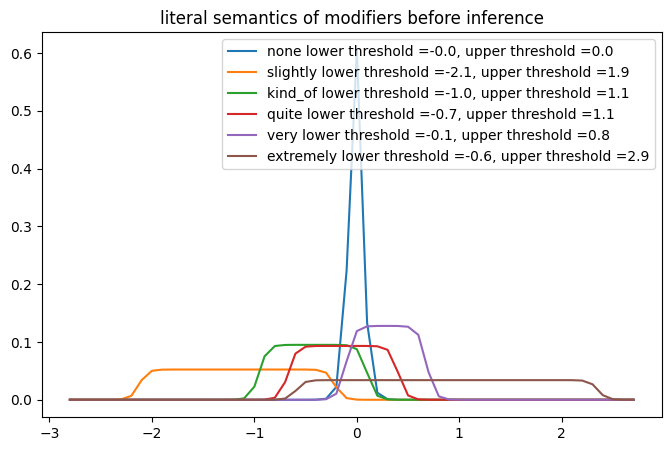}
    \caption{Visualization of Example Literal Semantics}
    \label{fig:literal_semantics}
\end{figure}

It is a topic of debate whether the literal semantics of the weaker form of utterances (e.g. 'kind of cold' instead of 'very cold') falsifies the stronger interpretation, or whether this is due to scalar implicatures. We model the literal meaning using a double threshold, thus allowing the possibility to model weak utterances as inconsistent with the stronger meaning (example shown in figure \ref{fig:literal_semantics}). The flexibility with double thresholding leaves it up to the model to decide whether the fact that 'slightly cold' is not used to describe stronger states is due to it's semantics, or whether it is due to pragmatics (scalar implicatures).
Furthermore, while the literal meaning of a modifier is conventionally modeled by a discrete threshold, we adopt a smooth threshold. As we adopted gradient optimization to find the best fitting parameters, smooth thresholding was necessary to carry out gradient descent effectively.

\subsubsection{Robustness check: Fitting model to narrator data}
By collecting participant judgements in scenarios where the narrator was making an utterance, our narrator scenarios aimed to measure interpretation of modifiers in the absence of social considerations. Therefore, under the assumption that the only difference between the narrator and dialogue scenario is the presence of the social utility term. This means that theoretically, the model fit on the dialogue data except with the social utility term adjusted to zero should fit the narrator data well.
To test this, we ran optimization to find the best fitting model to the narrator data, where the model family is the same as the one for fitting dialogue data, except social utility is 0 for all utterances (we call this model the 'narrator model'). We compared this with the best performing model for the dialogue data (which we call the 'dialogue model') and computed the logloss of this model on the narrator data (where I replaced the soc term of the original model with 0).

\begin{table}[b!]
\centering
\small
\caption{logloss of model on the narrator data}
\label{tab:logloss_narrator}
\begin{tabular}{@{}p{0.5\linewidth}|p{0.12\linewidth}|p{0.12\linewidth}|p{0.12\linewidth}@{}}
\hline
\textbf{model type} & \textbf{UK} & \textbf{US} & \textbf{Both} \\ \hline
\textbf{narrator model}&4488&4792& 9303\\ \hline
\textbf{dialogue model  with soc = 0} & 4528 & 4821 & 9328\\
\hline
\end{tabular}
\end{table}

The log loss increased by a significant amount (about 30), suggesting that the pragmatic considerations captured by $U_{soc} $ are different from the differences in pragmatic considerations between the narrator and dialogue condition.
\subsubsection{Robustness to dropping modifiers}
Our experiment measured responses of modifier interpretation in the context of modifying seven different predicates. By aggregating data on modifier interpretations from a varied set of predicates and social contexts, we aimed to understand the pure effect of modifier rather than their effect in a specific context or predicate it is modifying. To see if our results were robust to the choice of modifiers, we fit the model with a subset of the data by dropping all measurements associated with a chosen predicate. We found that when dropping predicates 'difficult' or 'impressive' the best fitting parameters for this modified data still achieved similar log losses when used to predict the original data (i.e. adding back the dropped out data). However, dropping 'extremely' had quite a large difference on the overall logloss, suggesting that the modeling may have been overly affected by specific scenarios in the behavioral experiment.
\begin{table}[b!]
\centering
\small
\caption{logloss of parameters on the original data}
\label{tab:logloss_original}
\begin{tabular}{@{}p{0.4\linewidth}|p{0.15\linewidth}|p{0.15\linewidth}|p{0.15\linewidth}@{}}
\hline
\textbf{data the model was optimized for} & \textbf{UK} & \textbf{US} & \textbf{Both} \\ \hline
\textbf{original data (baseline)}& 5070&6124& 11250 \\ \hline
\textbf{data without 'extremely'} & 5082 & 6139 & 11287\\ \hline
\textbf{data without 'impressive'} & 5074 & 6127 & 11258\\ \hline
\textbf{data without 'difficult'}& 5072 & 6125 & 11251\\ \hline
\end{tabular}
\end{table}

\end{document}